\documentclass[10pt,twocolumn,letterpaper]{article}

\usepackage[pagenumbers]{iccv} % To force page numbers, e.g. for an arXiv version

\newcommand\textred[1]{\textcolor{red}{#1}}
\newcommand\textblue[1]{\textcolor{blue}{#1}}

\definecolor{cvprblue}{rgb}{0.21,0.49,0.74}
\usepackage[pagebackref,breaklinks,colorlinks,citecolor=cvprblue]{hyperref}
\usepackage{amssymb}
\usepackage{amsmath}
\usepackage{graphicx}
\usepackage[ruled,vlined,linesnumbered]{algorithm2e}
\usepackage{algpseudocode}
\usepackage{multirow}
\def\paperID{14059} % *** Enter the Paper ID here
\def\confName{ICCV}
\def\confYear{2025}

\title{Block-based Symmetric Pruning and Fusion for Efficient Vision Transformers}

\author{Yi-Kuan Hsieh, Jun-Wei Hsieh, Xin Li, Yu-Ming Chang, Yu-Chee Tseng\\
College of Artificial Intelligence and Green Energy\\
National Yang Ming Chiao Tung University, Hsinchu, Taiwan\\
Computer Science Department \\
University at Albany, SUNY, NY, USA\\
Computer Science Information Engineering Department \\
University at Albany, SUNY, NY, USA\\
}

\begin{document}
\maketitle

%%%%%%%%%%%%%%%%%%%%%%%%%%%%%%%%%
%%%%%%%%%%%%%%%%%%%%%%%%%%%%%%%%%
\begin{abstract}

Vision Transformer (ViT) has achieved impressive results across various vision tasks, yet its high computational cost limits practical applications. Recent methods have aimed to reduce ViT's $O(n^2)$ complexity by pruning unimportant tokens. However, these techniques often sacrifice accuracy by independently pruning query (Q) and key (K) tokens, leading to performance degradation due to overlooked token interactions. To address this limitation, we introduce a novel {\bf Block-based Symmetric Pruning and Fusion} for efficient ViT (BSPF-ViT) that optimizes the pruning of Q/K tokens jointly. Unlike previous methods that consider only a single direction, our approach evaluates each token and its neighbors to decide which tokens to retain by taking token interaction into account. The retained tokens are compressed through a similarity fusion step, preserving key information while reducing computational costs.
The shared weights of Q/K tokens create a symmetric attention matrix, allowing pruning only the upper triangular part for speed up. BSPF-ViT consistently outperforms state-of-the-art ViT methods at all pruning levels, increasing ImageNet classification accuracy by 1.3\% on DeiT-T and 2.0\% on DeiT-S, while reducing computational overhead by 50\%. It achieves 40\% speedup with improved accuracy across various ViTs.

\end{abstract}
%%%%%%%%%%%%%%%%%%%%%%%%%%%%%%%%%
%%%%%%%%%%%%%%%%%%%%%%%%%%%%%%%%%

%--------------------------------------
\begin{figure}[t]
\centerline{
  \includegraphics[width=\linewidth]{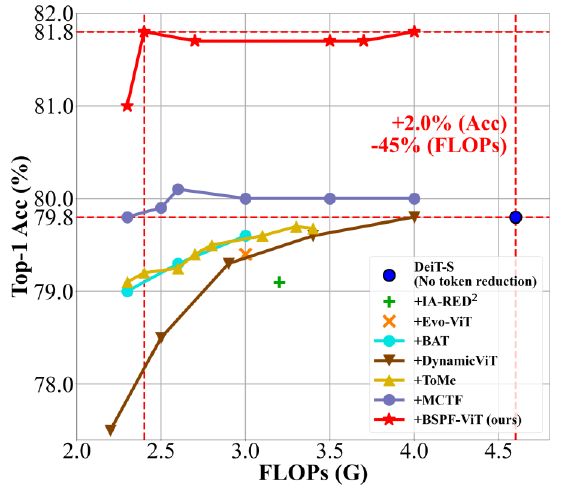}
  \vspace{-2mm}
}
\caption{\textbf{ViT token reduction performance comparison:} 
\textcolor{blue}{Blue} circle represents the DeiT-S~\cite{touvron2021training} base model. While previous token reduction methods improve speed, they compromise accuracy and model effectiveness. The proposed BSPF-ViT (\textcolor{red}{red}) effectively reduces complexity while enhancing performance.}
\label{fig:Line chart}
\vspace{-4mm}
\end{figure}
%--------------------------------------

%%%%%%%%%%%%%%%%%%%%%%%%%%%%%%%%%%%%%%%
%%%%%%%%%%%%%%%%%%%%%%%%%%%%%%%%%%%%%%%
%%%%%%%%%%%%%%%%%%%%%%%%%%%%%%%%%%%%%%%
\section{Introduction}

Vision Transformer (ViT) \cite{dosovitskiy2020image} was introduced to handle vision tasks using {\em self-attention} and the {\em attention mechanism}, originally designed to assess the importance of individual tokens in a sequence. Since then, the Transformer has become the dominant architecture across a wide range of vision tasks, including classification\cite{dosovitskiy2020image,liu2021swin,touvron2021training,touvron2021going,wang2021pyramid}, segmentation~\cite{liu2021swin,strudel2021segmenter,xie2021segformer}, and object detection~\cite{carion2020end,wang2021pyramid,zhu2020deformable}.  ViTs, relying solely on self-attention and MLP layers, provide remarkable flexibility and performance advantages over traditional methods like convolutional neural networks (CNNs). However, the quadratic computational complexity of self-attention with respect to the number of tokens presents a major challenge, especially as interest grows in large-scale foundation models such as CLIP~\cite{radford2021learning} and Llama2~\cite{touvron2023llama}. To mitigate this, several studies~\cite{beltagy2020longformer, kitaev2020reformer, wang2020linformer, ma2022mega} have proposed efficient self-attention mechanisms, including local self-attention within predefined windows~\cite{chu2021twins,LinICCV2023,liu2021swin,Ali2023learning, Dai2024CVPR,Fan2024CVPR}.

There is growing interest in token reduction methods to optimize ViTs without changing their architecture. These methods fall into three main categories: \textbf{Token pruning}~\cite{fayyaz2022adaptive, meng2022adavit, pan2021ia, rao2021dynamicvit, yin2022vit}, which removes uninformative tokens to reduce the total count; \textbf{Token reorganization}~\cite{bolya2022token, kong2022spvit, liang2022not, long2023beyond, marin2023token}, which fuses tokens instead of discarding them; and \textbf{Token squeezing} \cite{Wei_2023_CVPR, lee2024multi}, which minimizes message loss without creating extra representative tokens.

These token pruning methods reduce the number of tokens, thereby lowering the computational cost of performing the dot product between query and key elements. However, most fusion methods tend to degrade model performance and usually focus on pruning in only one direction (along query or key). Our observations indicate that, while these methods effectively reduce computational costs, they often eliminate critical information in the attention map and typically base pruning decisions solely on individual tokens.
%(as in Fig.~\ref{fig:attention map}b).

% These token pruning methods reduce the token count from  $N$ to $C$, decreasing self-attention complexity of computing the dot product of query $Q$ and key $K$ from $O(N^2)$ to $O(C^2)$. However, most fusion methods often degrade performance and focus on pruning in a single direction (along query or key). Our observations show that while these methods reduce computational costs, they often prune essential information in the attention map and typically base pruning decisions only on individual tokens.

%\textcolor{blue}{Despite these efforts, performance degradation is commonly observed in most token-fusion methods, and all methods primarily focus on one-dimensional token pruning. Our observations indicate that although existing methods effectively reduce computational complexity, they often inadvertently prune important information in the attention map. Furthermore, these methods typically consider only the information represented by a single token when making one-dimensional token pruning decisions. Our experiments demonstrate that considering the information content of adjacent tokens more than 1-D directions yields significantly better results than evaluating single tokens in isolation.} 

\begin{figure}[t]
\centerline{
  \includegraphics[width=\linewidth]{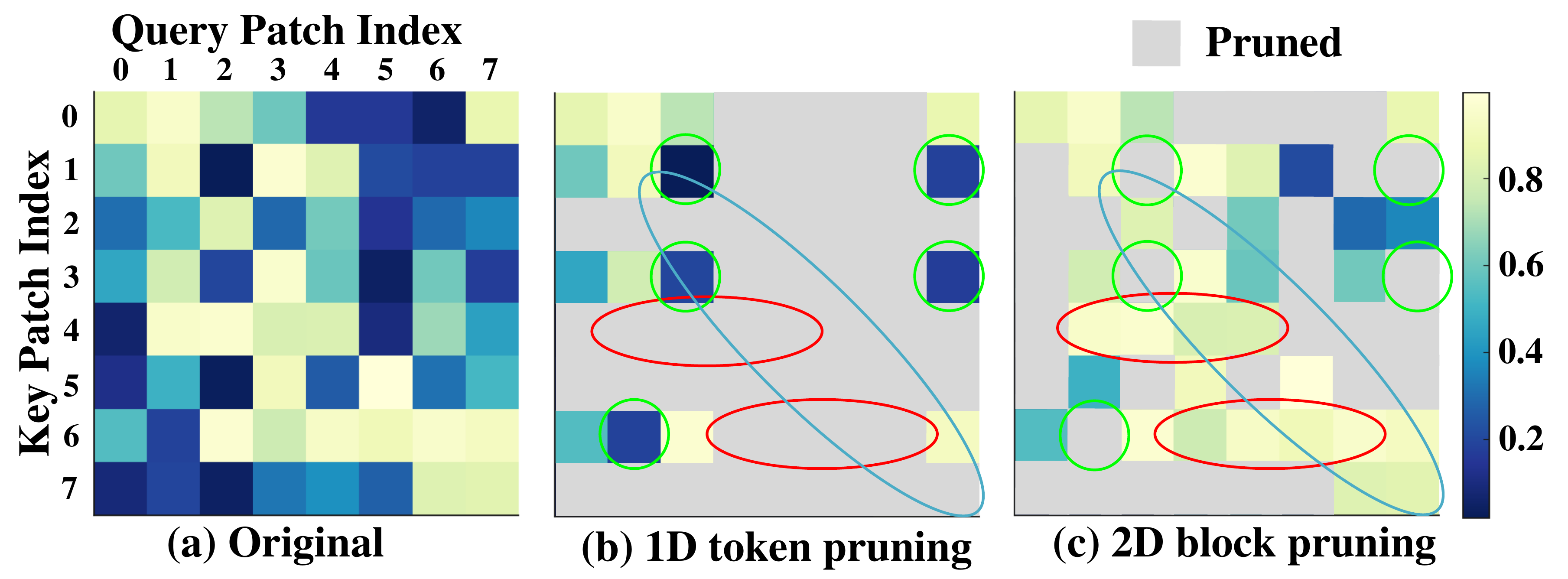} 
  \vspace{-2mm}
}
\caption{
{\bf Comparison of ViT token reduction strategies:} 
(a) The original attention map. 
(b) Result after 1D token pruning, where important information along the main diagonal is discarded (\textblue{blue} ovals). 
(c) Our proposed 2D block pruning effectively preserves critical information, highlighted in \textred{red} and \textblue{blue} ovals.
}
\label{fig:attention map}
\vspace{-4mm}
\end{figure}

To address these issues, we propose a novel {\bf block-based symmetric pruning and fusion} for efficient ViT (BSPF-ViT) method that optimizes Visual Transformers by applying {\em 2D block-based pruning} and then fusing tokens based on similarity. Unlike previous methods that focus on individual tokens, this block-based pruning accounts for token attention from both the query and key directions. It performs a comprehensive importance assessment of adjacent tokens within a block before deciding which tokens to prune. To preserve key information, we incorporate pruned tokens into the retained ones through a similarity-based weighted fusion scheme. To further simplify self-attention computation, we use shared weights for the query and key in ViT, making the attention matrix {\em symmetric}. This symmetry allows us to prune only the upper triangular portion of the matrix and map the pruned locations to the lower triangular part, which improves efficiency.  Fig.~\ref{fig:attention map} provides a high-level comparison of ViT token reduction strategies, and Fig.~\ref{fig:pruning} overviews the proposed VTPS-ViT method.
  
Our experiments show that  considering each token jointly with neighboring tokens leads to significantly better outcomes. Our approach outperforms existing methods in both performance and computational efficiency, as shown in Fig.~\ref{fig:Line chart}. Our method increases ImageNet classification accuracy by 1.7\% on DeiT-T and 2.0\% on DeiT-S, while reducing computational overhead in FLOPS by approximately 45\%. Similar speedups (40\%) and performance gains (0.9\% and 0.8\%) are observed in T2T-ViT~\cite{yuan2021tokens} and LV-ViT~\cite{jiang2021all}, respectively.
Our contributions are summarized as follows:
\begin{itemize}[leftmargin=10pt] \itemsep -.1em

\item We introduce a novel block-based pruning method for accelerating ViTs by assessing token importance in both the query (Q) and key (K) directions. This 2D pruning strategy sets our approach apart from conventional 1D pruning, which considers only the query or key direction.

\item We use a similarity-based fusion method to incorporate information from pruned tokens into retained tokens, preserving essential information and enhancing accuracy across various tasks.

\item We use shared weights for the query and key in ViT to make the attention matrix symmetric, enabling more efficient token pruning.

\item Our method surpasses existing ViT token pruning approaches in performance, while significantly reducing computational complexity. 

%Specifically: Achieves gains of 0.9\% and 1.4\% in DeiT-T and DeiT-S, respectively, while reducing the FLOP count by approximately 45\%. Observes similar speedups (40\%) in T2T-ViT and LV-ViT, resulting in performance gains of 0.9\% and 0.8\%, respectively.
\end{itemize}

%--------------------------------------
%--------------------------------------

%%%%%%%%%%%%%%%%%%%%%%%%%%%%%%%%%
%%%%%%%%%%%%%%%%%%%%%%%%%%%%%%%%%
%%%%%%%%%%%%%%%%%%%%%%%%%%%%%%%%%
\section{Related Work}

%%%%%%%%%%%%%%%%%%%%%%%%%%%%%%%%%
%%%%%%%%%%%%%%%%%%%%%%%%%%%%%%%%%
\subsection{Vision Transformers}

Vision Transformers (ViTs)~\cite{dosovitskiy2020image} were introduced to handle vision tasks, and advancements like DeiT~\cite{touvron2021training} and CaiT~\cite{touvron2021going} have improved their data efficiency and scalability. Recent studies~\cite{Chen_2021_ICCV, Dong_2022_CVPR, Heo_2021_ICCV, liu2021swin, wang2021pyramid} have incorporated CNN-like inductive biases into ViTs, such as locality and pyramid structures. Meanwhile, other research~\cite{bachmann2022multimae, he2022masked, touvron2021going, wu2022denoising, zhai2022scaling} has focused on scaling and self-supervised learning to improve vanilla ViTs. Despite these advancements, the quadratic complexity of ViTs still poses a significant challenge for model scaling.

To address this complexity, Reformer~\cite{kitaev2020reformer} reduces the quadratic complexity to $O(N \log N)$ using hashing functions, while Linformer~\cite{wang2020linformer}, Performer~\cite{choromanski2020rethinking}, and Nystromformer~\cite{xiong2021nystromformer} achieve linear complexity with approximated linear attention. Furthermore, several studies~\cite{arar2022learned, chu2021twins, Dong_2022_CVPR, liu2021swin} have explored sparse attention mechanisms by reducing the number of keys or queries. Swin Transformer~\cite{liu2021swin} and Twins~\cite{chu2021twins} mitigate complexity further by applying local attention within fixed-size windows.

%--------------------------------------
\begin{figure*}[t]
\centerline{
  \includegraphics[width=0.9\linewidth]{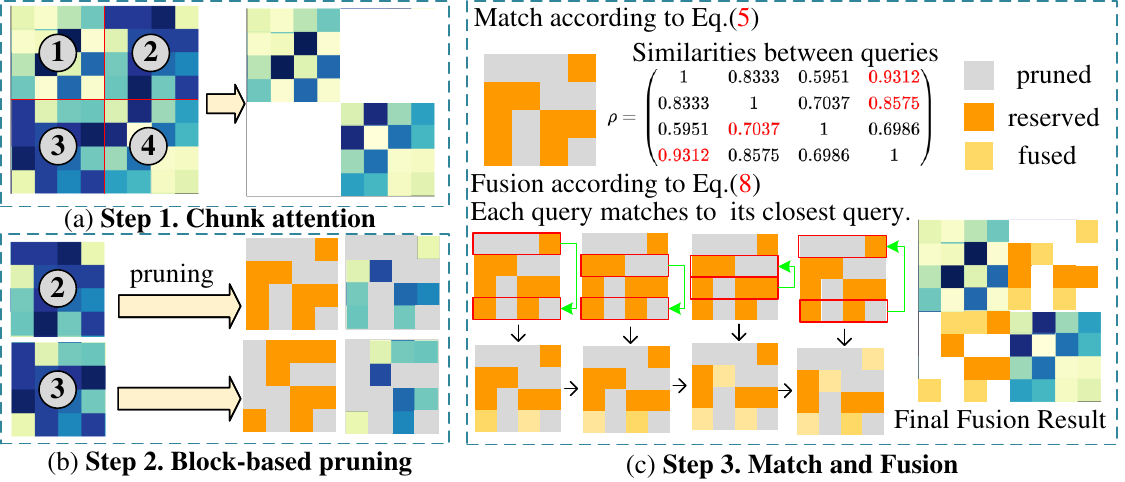}   
  \vspace{-2mm}
}
\caption{ 
The proposed {\bf block-based symmetric pruning and fusion} for efficient ViT (BSPF-ViT) consists of three main steps: (a) {\em Chunk-based attention} ($\S$~\ref{sec:chunk:attention}), which is modified from the traditional self-attention mechanism. 
(b) {\em Block-based pruning} ($\S$~\ref{sec:block:pruning}) is applied to the sparse regions, targeting sparse regions to determine which elements to retain or prune. 
(c) {\em Matching and fusion} ($\S$~\ref{sec:match:fusion}), where each row of the attention matrix is matched against other rows to determine the most similar row for fusion.
}
\label{fig:pruning}
\vspace{-4mm}
\end{figure*}
%--------------------------------------
%%%%%%%%%%%%%%%%%%%%%%%%%%%%%%%%%
%%%%%%%%%%%%%%%%%%%%%%%%%%%%%%%%%
\subsection{Self Attention}

Recent advancements in ViTs have significantly improved performance across various computer vision tasks. Several studies have introduced novel architectures to enhance both efficiency and performance. For instance, GCViT~\cite{Ali2023learning} combines global and local self-attention to effectively model both long- and short-range spatial interactions. The SMT~\cite{LinICCV2023} integrates convolutional networks with ViTs, featuring multi-head and scalre-aware modules. The SHViT~\cite{yun2024shvit} reduces computational redundancy through a larger-stride patchify stem and single-head attention. The RMT~\cite{Fan2024CVPR} introduces explicit spatial priors via a spatial decay matrix based on Manhattan distance. TransNeXt~\cite{Dai2024CVPR} addresses depth degradation and enhances local modeling by using Aggregated Attention, a biomimetic token mixing design, and Convolutional GLU as a channel mixer. These architectures have achieved state-of-the-art performance on tasks such as ImageNet classification, COCO object detection, and ADE20K semantic segmentation.

% Recent advancements in ViTs have significantly improved performance across various computer vision tasks. Several studies have introduced novel architectures to enhance both efficiency and performance. For instance, the Global Context ViT (GC ViT)~\cite{Ali2023learning} combines global and local self-attention to effectively model both long- and short-range spatial interactions. The Scale-Aware Modulation Transformer (SMT)~\cite{LinICCV2023} integrates convolutional networks with ViTs, introducing modules like Multi-Head Mixed Convolution and Scale-Aware Aggregation. The Single-Head Vision Transformer (SHViT)~\cite{yun2024shvit} reduces computational redundancy through a larger-stride patchify stem and single-head attention. The Retentive Network-inspired Transformer (RMT)~\cite{Fan2024CVPR} introduces explicit spatial priors via a spatial decay matrix based on Manhattan distance. TransNeXt~\cite{Dai2024CVPR} addresses depth degradation and enhances local modeling by using Aggregated Attention, a biomimetic token mixing design, and Convolutional GLU as a channel mixer. These architectures have achieved state-of-the-art performance on tasks such as ImageNet classification, COCO object detection, and ADE20K semantic segmentation.

%%%%%%%%%%%%%%%%%%%%%%%%%%%%%%%%%
%%%%%%%%%%%%%%%%%%%%%%%%%%%%%%%%%
\subsection{Token Reduction in Vision Transformers}

The primary computational burden in ViTs arises from self-attention mechanisms. Recent studies~\cite{bolya2022token, fayyaz2022adaptive, kong2022spvit, liang2022not, long2023beyond, marin2023token, meng2022adavit, pan2021ia, rao2021dynamicvit, yin2022vit} have focused on alleviating this burden by reducing the number of tokens, typically by assessing the importance of each token and discarding those considered less essential.
Token reduction methods can be broadly categorized into three categories: \textbf{token pruning}, \textbf{token reorganization}, and \textbf{token squeezing}, each with unique methods to achieve efficient token reduction.

% {\bf Token pruning} involves discarding less significant tokens based on criteria such as attention scores or relevance within the overall context. For example, Fayyaz {\em et al.}~\cite{fayyaz2022adaptive} introduced a parameter-free Adaptive Token Sampler (ATS) module, halving GFLOPs without compromising accuracy. AdaViT~\cite{meng2022adavit} improves efficiency by selectively retaining valuable patches, attention heads, and transformer blocks, achieving 2$\times$ speedup with minimal accuracy loss. IA-RED$^2$~\cite{pan2021ia} uses an interpretable framework to dynamically eliminate uncorrelated tokens, yielding up to 1.4$\times$ speed increase. DynamicViT~\cite{rao2021dynamicvit} progressively prunes redundant tokens, reducing FLOPs by 31\% and improving throughput by over 40\%. A-ViT~\cite{yin2022vit} speeds up inference by adaptively stopping token computation without additional parameters or calculations. While this approach enhances processing speed, it can sometimes lead to the loss of critical information, potentially affecting performance. 
% The Hessian-aware token pruning in \cite{yang2023global} is a global  approach that reassigns parameters across transformer blocks, challenging the standard uniform dimension design in ViTs. However, this Hessian-based pruning requires significant computational resources and exploring the entire ViT model design space increases the complexity of design and implementation.

{\bf Token pruning} discards less significant tokens based on criteria like attention scores. For instance, Fayyaz’s {\em et al.}~\cite{fayyaz2022adaptive} proposes an Adaptive Token Sampler (ATS) module to reduce GFLOPs by half without accuracy loss, while AdaViT~\cite{meng2022adavit} retains key patches and blocks, doubling speed with minimal impact. IA-RED$^2$~\cite{pan2021ia} dynamically removes uncorrelated tokens, achieving up to 1.4$\times$ speedup. DynamicViT~\cite{rao2021dynamicvit} progressively prunes redundant tokens, cutting FLOPs by 31\% and boosting throughput by 40\%, and A-ViT~\cite{yin2022vit} adaptively stops token computation for faster inference. Though efficient, these methods can risk critical information loss. Hessian-aware pruning \cite{yang2023global} reallocates parameters globally across transformer blocks but requires significant computational resources, adding complexity.

% , which limits its feasibility in resource-constrained environments

{\bf Token reorganization} involves consolidating less significant tokens into a single and representative token. For instance, SPViT~\cite{kong2022spvit} uses a latency-aware soft token pruning framework to reduce computational costs while meeting device-specific latency requirements. Token Merging (ToMe)~\cite{bolya2022token} enhances throughput by combining similar tokens without the need for retraining. EViT~\cite{liang2022not} reorganizes tokens by retaining the most attentive tokens and merging the least attentive ones, improving both efficiency and accuracy. An efficient token decoupling and merging approach \cite{long2023beyond} uses token importance and diversity to separate attentive from inattentive tokens, enabling more effective pruning. Token pooling~\cite{marin2023token} treats tokens as samples from a continuous signal and selects the optimal set to approximate it, improving the computation-accuracy trade-off. These reorganization methods reduce computational costs while preserving important information.

% {\bf Token squeezing} involves pairing less significant tokens with more important ones based on their similarity, then fusing them together. For example, Multi-criteria Token Fusion (MCTF)~\cite{lee2024multi} merges tokens progressively based on criteria like similarity, informativeness, and the size of the fused tokens, while using one-step-ahead attention to assess token informativeness. Joint Token Pruning and Squeezing (TPS) \cite{Wei_2023_CVPR} combines token pruning with squeezing, preserving information from pruned tokens by incorporating them into the remaining ones. It aims to balance computational efficiency with accuracy preservation. However, all three methods prune tokens in only one direction.Token Fusion (ToFu)~\cite{kim2024token} combines token pruning and merging techniques, introducing MLERP merging to maintain feature norm distribution, resulting in improved computational efficiency and model accuracy for both classification and image generation tasks. Zero-TPrune~\cite{wang2024zero} offers a zero-shot token pruning method that leverages the attention graph of pre-trained Transformers, eliminating the need for fine-tuning while reducing FLOPs and improving throughput with minimal accuracy loss. ELSA~\cite{huang2024elsa} exploits layer-wise N:M sparsity for ViTs, considering both sparsity levels and expected throughput improvement on N:M sparsity-supporting accelerators. This approach achieves significant reductions in FLOPs for Swin-B and DeiT-B models with only marginal accuracy degradation on ImageNet.

{\bf Token squeezing} pairs less significant tokens with more important ones based on similarity, then fuses them. For instance, Multi-criteria Token Fusion (MCTF)~\cite{lee2024multi} progressively merges tokens using criteria like similarity, informativeness, and fused token size, with one-step-ahead attention to assess informativeness. Joint Token Pruning and Squeezing (TPS) \cite{Wei_2023_CVPR} integrates pruning and squeezing, retaining pruned token information to balance efficiency and accuracy. However, these methods prune tokens in one direction only. Token Fusion (ToFu)~\cite{kim2024token} combines pruning and merging via MLERP, preserving feature norms and enhancing efficiency and accuracy in classification and image generation. Zero-TPrune~\cite{wang2024zero} introduces zero-shot pruning, leveraging pre-trained Transformer attention graphs to reduce FLOPs and improve throughput without fine-tuning. ELSA~\cite{huang2024elsa} uses layer-wise N:M sparsity in ViTs to optimize sparsity for accelerators, significantly reducing FLOPs in Swin-B and DeiT-B models with minimal accuracy loss on ImageNet.

%While effective at reducing computation time in self-attention mechanisms of vision transformers, these approaches often lead to decreased accuracy. This paper considers that token pruning across multiple dimensions can potentially achieve a better balance between computational efficiency and model accuracy.

%%%%%%%%%%%%%%%%%%%%%%%%%%%%%%%%%
%%%%%%%%%%%%%%%%%%%%%%%%%%%%%%%%%

In summary, existing ViT token reduction approaches focus on discarding unimportant tokens along a single direction, either query or key.  As shown in Fig.~\ref{fig:attention map}, although sequence token pruning can effectively remove low-score tokens, it often discards important information (marked by red circles) along the main diagonal direction (the blue circles).  In addition, when some  query or key directions are selected, unimportant entries with lower attention values along the selected directions are still kept (the green circles in Fig.~\ref{fig:attention map}(b) and Fig.~\ref{fig:attention map}(c)). In this paper, we propose a remedy to perform block-based pruning to keep entries with higher attentions by filtering out unimportant tokens.

%%%%%%%%%%%%%%%%%%%%%%%%%%%%%%%%%
%%%%%%%%%%%%%%%%%%%%%%%%%%%%%%%%%
%%%%%%%%%%%%%%%%%%%%%%%%%%%%%%%%%
\section{Method}
\label{sec:method}

The proposed {\bf block-based symmetric pruning and fusion} (BSPF-ViT) enhances ViT efficiency through a streamlined process outlined in Fig.~\ref{fig:pruning}, consisting of three main steps: 
(1) {\em Chunk-based attention:} This adapts traditional self-attention by dividing input tokens into chunks and applying self-attention within each chunk ($\S$~\ref{sec:chunk:attention}). 
(2) {\em Block-based pruning:} Building on chunk-based attention, this approach introduces a block-based pruning strategy that retains only the most important entries in each block. This reduces computational load while preserving essential information and is applied to sparse regions to selectively retain or prune elements based on sparsity ($\S$~\ref{sec:block:pruning}).
(3) {\em Matching and fusion:} Each row of the attention matrix is compared with others to identify and fuse it with the most similar row. By aligning pruned tokens with their closest matches, the model preserves critical information flow and minimizes data loss.($\S$~\ref{sec:match:fusion}).

% %--------------------------------------
% \begin{figure}[h]
%     \includegraphics[scale = 0.24]{tradition_transformer.pdf}   
%     \caption{ Traditional Transformer Architecture. The input tokens $x=\left[ x_{1},x_{2},\cdots,x_{n} \right]$ include $n$ tokens which are from input image. Each token vector $x_{i}$ has $d$ as channel dimension. The regular transformer calculates the attention score for each pair of token vectors $x_i$ and $x_j$.}
%     \label{fig:tradition traansformer}
% \end{figure}
% %--------------------------------------

%Details of each step are addressed as follows.

% and applies block pruning in the upper and lower triangles of the matrix (see Fig. \ref{fig:attention map}(b)). Additionally, we merge the pruned sections back into nearby significant areas based on their similarity, thereby retaining essential information more effectively.  This 2D block pruning approach can enhance the efficiency and accuracy of vision transformers by preserving more valuable information while still reducing computational load.

%--------------------------------------
\begin{figure}[t]
\centerline{
  \includegraphics[width=\linewidth]{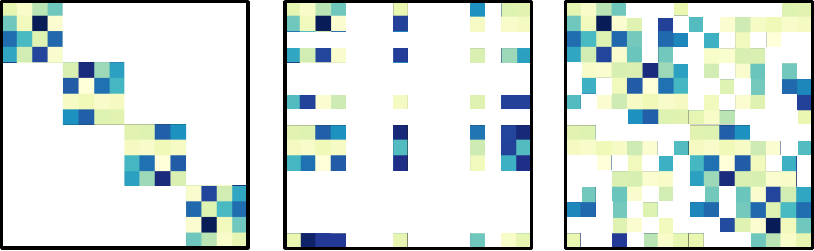}
} 
\centerline{
  \small{ (a) Chunk attention (b) Pruning attention (c) Block-based attention }
  \vspace{-2mm}
}
\caption{ (a) Chunk attention only considers self attention within chuck. (b) Pruning attention takes into account some features that are far away from itself, but it is still sparse. (c) Block-based pruning combines the advantages of chunk attention and pruning attention, and can have richer features.}
\label{fig:block_pruning} 
\vspace{-4mm}
\end{figure}
%--------------------------------------

%%%%%%%%%%%%%%%%%%%%%%%%%%%%%%%%%
%%%%%%%%%%%%%%%%%%%%%%%%%%%%%%%%%
\subsection{Chunk-based Attention}
\label{sec:chunk:attention}

For a set of $N$ tokens $\mathcal{T}=\{T_i \; | \; i=1,....,N\}, T_i \in R^d$, the self-attention~\cite{vaswani2017attention} of $\mathcal{T}$, denoted  $\mathrm{attn}()$, yields an attention matrix $\textbf{A}$, where $d$ is the feature dimension. Let $\mathcal{Q}$ and $\mathcal{K}$ denote the corresponding query and key set of $\mathcal{T}$: $\mathcal{Q} = \{Q_i\}$ and $\mathcal{K} = \{K_j\}$ for $T_i$ and $T_j$, respectively. $\textbf{A}(i,j)$, the $(i,j)$ entry of $\textbf{A}$, is calculated as:
\begin{equation}
   \textbf{A}(i,j) = \mathrm{attn}(Q_i,K_j)=\mathrm{softmax}\left(\frac{Q_i^T K_j}{\sqrt{d}}\right).
\label{eq:Attention}
\end{equation}
Despite the outstanding expressive power of self-attention, it does not computationally scale well to large $N$ due to its quadratic time complexity $O(N^2)$. To address this problem, prior works~\cite{ma2022mega, ju2021chunkformer, zha2021shifted} divide the input tokens into $C$ number of chunks according to the chunk size $\Omega$, where each chunk performs its own self attention mechanism.  The chunk-based attention is to divide the input tokens $\mathcal{T}$ into $C$ chunks: $\mathcal{T}=\{\textbf{T}_m\}_{m=1,...,C}$. Each chunk $\textbf{T}_m$ has $\Omega$ tokens: $\Omega=N/C$, where $\textbf{T}_m$=$\{T_{m_1}, ...,T_{m_\Omega}\}$.  The concept of chunk-based attention applies the attention operation individually to each chunk so that the time complexity of attention ranges from $O(N^2)$ to $O(C \Omega^2)=O(N\Omega)$, where $\Omega<<N$. Although the chunk-based transformer greatly reduces the computational complexity, there is still a fatal problem; that is,  they do not consider the relationship between distant tokens among different tokens. This paper includes a new block-based pruning method to deal with the problems of chunk-based attention.

%--------------------------------------
\begin{figure}[t]
\centerline{
    \includegraphics[width=0.8\linewidth]{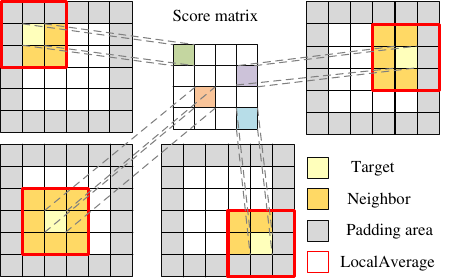}     \vspace{-2mm}
}    
\caption{This figure shows our proposed new score matrix that takes into account nearby information. }
\label{fig:neighbor}
\vspace{-4mm}
\end{figure}
%--------------------------------------

%%%%%%%%%%%%%%%%%%%%%%%%%%%%%%%%%
%%%%%%%%%%%%%%%%%%%%%%%%%%%%%%%%%
\subsection{Block-based Pruning}
\label{sec:block:pruning}

As shown in Fig.~\ref{fig:block_pruning}(a), the chunk-based attention method calculates only the entries along the diagonal band of the attention matrix $\textbf{A}$.  Other entries in the upper and lower matrix of $\textbf{A}$ are empty.  The second step of our method fills these empty entries through a block-based pruning method.  Similarly to chunk-based attention, we divide $\mathcal{Q}$ and $\mathcal{K}$ into $C$ chunks, respectively; that is, $\mathcal{Q}=\{\textbf{Q}_m \; | \; m=1,...,C\}$ and $\mathcal{K}=\{\textbf{K}_n \; | \; n=1,...,C\}$. The numbers of elements in $\textbf{Q}_m$ and $\textbf{K}_n$ are the same, \emph{i.e.}, $\Omega$, where  $\textbf{Q}_m = \{Q_{m1}, ...,Q_{m\Omega}\}$ and $\textbf{K}_n = \{K_{n1}, ...,K_{n\Omega}\}$.  Based on $\textbf{Q}_m$ and $\textbf{K}_n$, an attention block $\textbf{A}_{m,n}$ can be formed in the following:
\begin{equation}
    \textbf{A}_{m,n}(i,j) = \mathrm{attn}(Q_i,K_j),
\label{eq:attention}
\end{equation}
where $Q_i \in \textbf{Q}_m$ and $K_j \in \textbf{K}_n$.
Then, there are $\Omega\times \Omega$ entries in each block $\textbf{A}_{m,n}$.  For a block $\textbf{A}_{m,n}$, the block-based pruning strategy is to prune out its 50\% entries whose values are lower.  The ratio for pruning is decided on the trade-off between accuracy and efficiency (see the Ablation study in the experimental section).  In fact, ViT extracts tokens by dividing an image into different patches. After observation, directly pruning the tokens based on $\textbf{A}_{m,n}(i,j)$ will cause important noisy tokens to be removed. There exists high similarity between two adjacent patches due to the nature of the images.  This paper applies a learnable $3\times3$  convolution to locally weight $\textbf{A}_{m,n}(i,j)$ to obtain its smoothed score matrix $\textbf{S}_{m,n}(i,j)$.  As shown in Fig.~\ref{fig:neighbor}, for each token in a chunk, its importance is determined by considering its adjacent attention values rather than its single one.  This weighted scheme can reduce the effect of noisy tokens to get better accuracy but also increase the time complexity.  In $\S$~\ref{sec:Symmetry}, the concept of symmetry on attention matrix will be introduced to gain better efficiency.    
Before block pruning, a padding layer with values set to zero is added around each chunk attention matrix $\textbf{A}_{m,n}$. We apply a learnable $3 \times 3$ convolution kernel with stride 1 to smooth $\textbf{A}_{m,n}$ so that the score matrix $S_{m,n}(i,j)$ is obtained by $3 \times$ convolution:
\begin{equation}
    {\textbf{S}_{m,n}}(i,j) = \sum\limits_{k,l =  - 1}^1 {{w_{k, l}}{\textbf{A}_{m,n}}(i - k,j - l)},
\label{eq:ScoreMatrix}
\end{equation}
where $w_{i,j}$ is a learnable weight and $\sum\limits_{k,l =  - 1}^1 w_{k,l} $=1.  Since $\textbf{S}_{m,n}$ is decided by  taking into account the influence of surrounding neighboring tokens, it performs better than $A_{m,n}$ in our block-based token pruning.  Our block-based pruning strategy is to remove 50\% entries whose values in $\textbf{S}_{m,n}$ are lower.  With $\textbf{S}_{m,n}$, $\textbf{A}_{m,n}$ can be separated into a pruned attention matrix $\textbf{A}_{m,n}^p$ and a reserved attention matrix $\textbf{A}_{m,n}^r$ as:
\begin{equation}
\begin{aligned}
    \text{if } S_{m,n}(i,j) \text{ is listed in Top 50\%}:   \qquad  \qquad  \qquad   \ \ \ &\\
    A^r_{m,n}(i,j) = A_{m,n}(i,j) \text{ and } A^p_{m,n}(i,j) = 0, \\
    \text{else}:  \qquad \qquad \qquad  \qquad  \qquad  \qquad  \qquad  \quad \quad  \quad \quad \ \ &\\ 
    A^r_{m,n}(i,j) = 0 \text{ and } A^p_{m,n}(i,j) = A_{m,n}(i,j).
\end{aligned}
\label{eq:pruningratio}
\end{equation}
 % $if \textbf{S}_{m,n}(i,j) $ is listed in Top 50\%,$ \\
 %  $\quad \quad \textbf{A}_{m,n}^r(i,j)= \textbf{A}_{m,n}(i,j)$ and $\textbf{A}_{m,n}^p(i,j)=0;$\\
 % \quad  else \\
 %  \quad \quad $\textbf{A}_{m,n}^r(i,j)= 0$ and $\textbf{A}_{m,n}^p(i,j)= \textbf{A}_{m,n}(i,j)$.

% The time complexity to obtain its media is higher than the one to obtain its mean.  Thus, this paper uses a mean-based method to remove its unimportant entries.  The effect of cutting ratio will be discussed in experimental section.  Let $\mu_{m,n}$ denote the mean of $\textbf{A}_{m,n}$. With $\mu_{m,n}$, we can separate $\textbf{A}_{m,n}$ into two matrices: the reserved attention matrix $\textbf{A}_{m,n}^r$ and the pruned attention matrix $\textbf{A}_{m,n}^p$.   If $\textbf{A}_{m,n}(i,j)>\mu_{m,n}$, $\textbf{A}_{m,n}^r(i,j)$= $\textbf{A}_{m,n}(i,j)$ and $\textbf{A}_{m,n}^p(i,j)$=0; otherwise, $\textbf{A}_{m,n}^r(i,j)$=0 and $\textbf{A}_{m,n}^p(i,j)$= $\textbf{A}_{m,n}(i,j)$. After block-based pruning, the next step is to perform the process of matching and fusion.   

%--------------------------------------
\begin{figure}[t]
    \includegraphics[width=\linewidth]{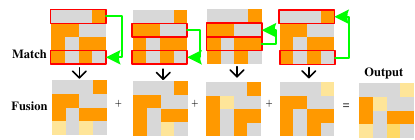}   
    \caption{ Each query finds its closest query and then contributes its information to the matched query for all important keys.}
    \label{fig:matching_fusion}
\end{figure}
%--------------------------------------
%%%%%%%%%%%%%%%%%%%%%%%%%%%%%%%%%
%%%%%%%%%%%%%%%%%%%%%%%%%%%%%%%%%
\subsection{Matching and Fusion}
\label{sec:match:fusion}

\begin{figure}[t]
\centerline{
  \includegraphics[width=\linewidth]{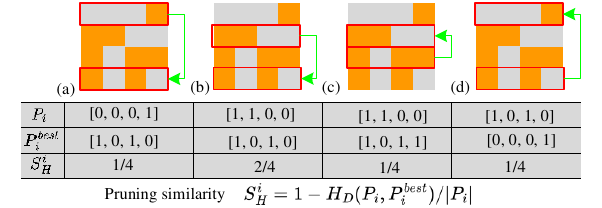}   
\vspace{-2mm}
}
 % \centerline{ \hspace{0.2cm} (a) \hspace{1.4cm} (b) \hspace{1.4cm} (c)\hspace{1.2cm} (d) }
\caption{An example calculating the pruning similarity.}
\label{fig:similarity_fusion}
\vspace{-4mm}
\end{figure}

After pruning, essential tokens in $\textbf{A}_{m,n}^p$ still contribute useful information to their best-matched tokens in $\textbf{A}_{m,n}^r$. In order to minimize data loss, we can align these tokens with their closest matches in $\textbf{A}_{m,n}^r$ to preserve the critical information flow. As shown in Fig.~\ref{fig:matching_fusion},
for each query $Q_i$ in $\textbf{Q}_m$, we find its best match $Q_{i}^{best}$ from $\textbf{Q}_m$ as follows:
\begin{equation}
   {Q_i^{best}} =  \mathop {\arg\max }\limits_{Q \in {Q_m} \& Q \neq Q_i} \frac{{{Q_i} \cdot Q}}{{||{Q_i}||||Q||}}.
\label{Query_Match}
\end{equation}
Let $i_b$ denote the query index of ${Q_i^{best}}$.
For each entry along the $i_b$-th row of $\textbf{A}_{m,n}^r$, the information of $\textbf{A}_{m,n}^p(i_b,j)$ is preserved to 
$\textbf{A}_{m,n}^r(i_b,j)$ as follows:
\begin{equation}
    \textbf{A}_{m,n}^r(i,j) =\textbf{A}_{m,n}^r(i,j) + \alpha \textbf{A}_{m,n}^p(i,j),
\label{eq:fusion}
\end{equation}
where $\alpha$ is a rate to determine how much information should be preserved from $\textbf{A}_{m,n}^p(i,j)$.  The value of $\alpha$ is set to be proportional to the cosine similarity $\rho_i^b$ between $Q_{i}$ and ${Q_i^{best}}$. Two tokens are similar not only from their visual features, but also from the pruned conditions associated with them.   In addition to the visual similarity between them, the similarity between their pruning conditions for keys is also incorporated into the calculation of $\alpha$.   As shown in Fig.~\ref{fig:similarity_fusion}, for queries $Q_i$ and $Q_i^{best}$ in case (a), their corresponding key positions to be removed are [0, 0, 0, 1] and [1, 0, 1, 0], where zero and one mean  the `pruned' and `reserved' status, respectively.  Let $P_i$ and $P_i^{best}$ denote these punning strings of $Q_i$ and $Q_i^{best}$, respectively. This paper uses the edit distance (or Hamming distance) to measure the distance between $P_i$ and $P_i^{best}$.  The more similar they are, the smaller the Hamming distance. Let $H_D(P_i,P_i^{best})$ denote the Hamming distance between $P_i$ and $P_i^{best}$.  Then, the similarity of pruning condition between $Q_i$ and $Q_i^{best}$ can be measured as:
\begin{equation}
   S_H^i = 1-H_D(P_i, P_i^{best})/|P_i|,
\label{Hamming similarity}
\end{equation}
where $|P_i|$ is the length of $P_i$. Then, Eq.(\ref{eq:fusion}) can rewritten as:  
\begin{equation}
   \textbf{A}_{m,n}^r(i_b,j) = \textbf{A}_{m,n}^r(i_b,j)+ \rho_i^b S_H^i \textbf{A}_{m,n}^p(i_b,j).
\label{simularity fusion}
\end{equation}
Fig.~\ref{fig:matching_fusion} shows the fusion process.  In real implementation, $\textbf{A}_{m,n}^r(i,j)$ is updated only if $\textbf{A}_{m,n}^r(i,j) > $ 0.  After that,  each row of $\textbf{A}_{m,n}^r$ is normalized with a SoftMax function.  Finally, we replace $\textbf{A}_{m,n}$ with $\textbf{A}_{m,n}^r$, \emph{i.e.}, $\textbf{A}_{m,n}$=$\textbf{A}_{m,n}^r$.  Algorithm~\ref{Algo:Algo1} describes the details of block-based pruning, matching, and fusion. 

%--------------------------------------
%--------------------------------------

%%%%%%%%%%%%%%%%%%%%%%%%%%%%%%%%%%%%%%%%%%%%%%%%%
\begin{algorithm}[t]
\caption{Block-based token pruning, matching, and fusion of the Transformer attention matrix}
\label{Algo:Algo1}
\SetAlgoLined
\textbf{Input:} query $\textbf{Q}_{m}$ and key $\textbf{K}_{n}$ matrices\\
\textbf{Output:} attention matrix $\textbf{A}_{m,n}$\\
For {$Q_i \in \textbf{Q}_m$ and $K_j \in \textbf{K}_n$, \textblue{// Pruning} }{\\
  \quad $A_{m,n}(i,j) =attention(Q_i,K_j)$} \\
 \quad // $S$: Scoring matrix \\
 \quad $\textbf{S}_{m,n}(i,j) = LocalAverage(A_{m,n}(i,j))$ according to Eq.\eqref{eq:ScoreMatrix}\\
  
For {$Q_i \in \textbf{Q}_m$ and $K_j \in \textbf{K}_n$, }{\\
  \quad  If $\textbf{S}_{m,n}(i,j) $ is in Top 50\%, \\
  \quad \quad $A_{m,n}^r(i,j)= A_{m,n}(i,j)$ and $A_{m,n}^p(i,j)=0$\\
 \quad  else \\
  \quad \quad $A_{m,n}^r(i,j)= 0$ and $A_{m,n}^p(i,j)= A_{m,n}(i,j)$\\
}
\quad Find $Q_i^{best}$ according to Eq.\eqref{Query_Match} \textblue{// Matching}\\
\quad  Let $i_{b}$  = the query index of ${Q_i^{b}}$\\
\quad  If $\textbf{A}_{m,n}^r(i_b,j)>0$, \textblue{// Fusion}\\  \quad \quad Update $\textbf{A}_{m,n}^r(i_b,j)$ according to Eq.\eqref{simularity fusion}
\\
Normalize each row of $\textbf{A}^r_{m,n}$ by SoftMax \\
$\textbf{A}_{m,n}=\textbf{A}_{m,n}^r$\\
\textbf{return} $\textbf{A}_{m,n}$

\end{algorithm}
%%%%%%%%%%%%%%%%%%%%%%%%%%%%%%%%%%%%%%%%%%%%%%%%%

%--------------------------------------
\begin{figure}[t]
\centerline{
    \includegraphics[width=\linewidth]{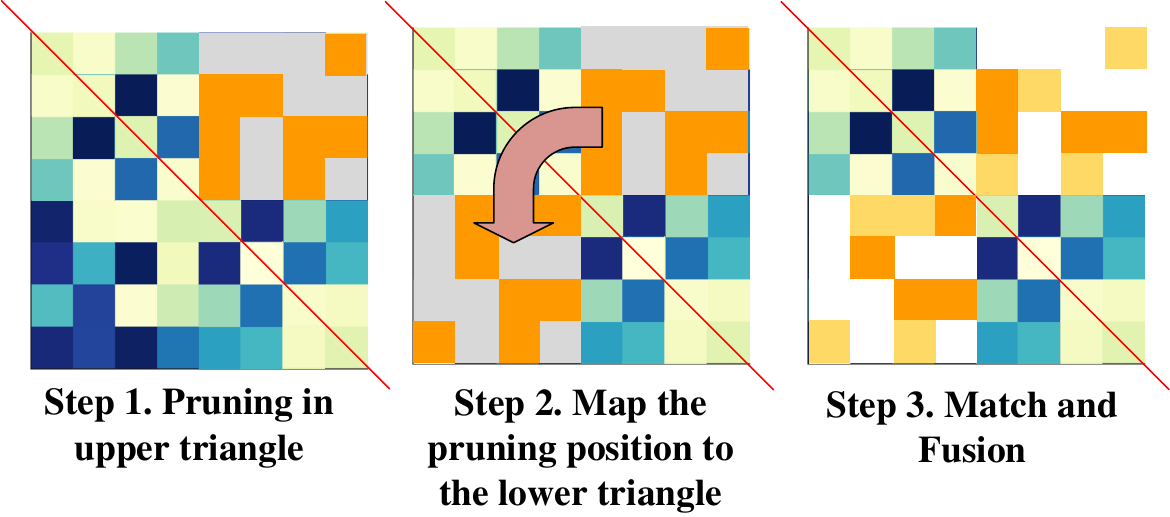} 
\vspace{-2mm}
}    
\caption{For a symmetric attention matrix, the pruning can be accelerated by pruning only the upper diagonal matrix, and then mirror to the lower half of the matrix. This effectively reduces the pruning cost and speed up the process. }
\label{fig:Symmetry}
\vspace{-4mm}
\end{figure}
%--------------------------------------

%%%%%%%%%%%%%%%%%%%%%%%%%%%%%%%%%
%%%%%%%%%%%%%%%%%%%%%%%%%%%%%%%%%
\subsection{Symmetry}
\label{sec:Symmetry}

In transformers, the Query, Key, and Value features are derived by segmenting the input text or image into tokens, represented as $X={x_1,x_2,\cdots,x_n}$. Each token is then transformed using separate weight matrices, $ W_q $, $ W_k $, and $ W_v $.
 As show in Fig.~\ref{fig:Symmetry} when the attention map is symmetric, pruning can be efficiently performed by focusing solely on the upper triangular portion of the matrix. The pruned positions can then be directly mapped to their corresponding locations in the lower triangular part. This design significantly reduces computational time.  One approach to achieve a symmetric attention matrix is by modifying  $A=Q^TK$ to $A=Q^TK+K^TQ$; however, this method is computationally expensive.
 
In natural language processing, it is intuitive for the influence of token $x_1$ on $x_2$ to be different from the influence of $x_2$ on $x_1$. However, in visual data, the influence of $x_1$ on $x_2$ is often equivalent to that of $x_2$ on $x_1$. Thus, we propose that in vision transformers, the query and key should be treated with equal significance by enforcing $W_k$ = $W_q$. This design ensures that the attention matrix  $A = Q^T K $ is inherently symmetric, eliminating the need for additional operations like $A=Q^TK+K^TQ$ to enforce symmetry.

\section{Experiment Results}

%%%%%%%%%%%%%%%%%%%%%%%%%%%%%%%%%%
%%%%%%%%%%%%%%%%%%%%%%%%%%%%%%%%%%
%\subsection{Main Results}

To validate the effectiveness of BSPF-ViT against state-of-the-art (SoTA) token reduction techniques, we evaluated a range of token pruning methods, including A-ViT~\cite{yin2022vit}, IA-RED2~\cite{pan2023interpretability}, DynamicViT~\cite{rao2021dynamicvit}, EvoViT~\cite{xu2022evo}, and ATS~\cite{fayyaz2022adaptive}, as well as token fusion methods such as SPViT~\cite{kong2022spvit}, EViT~\cite{liang2022not}, ToMe~\cite{bolya2022token}, BAT~\cite{long2023beyond}, ToFu~\cite{kim2024token}, Zero-TP~\cite{wang2024zero}, ELSA~\cite{huang2024elsa} and MCTF~\cite{lee2024multi}, using DeiT~\cite{touvron2021training} as the baseline. We report each method’s efficiency in GFLOPs and performance in Top-1 accuracy (\%). Furthermore, to evaluate our block-based symmetric pruning and fusion approach on other Vision Transformers, like T2T-ViT~\cite{yuan2021tokens} and LV-ViT~\cite{jiang2021all}, we compared our results to their respective official scores. In Tables~\ref{table: classification} to \ref{table: other transformer}, gray indicates the base model, while green and red represent performance improvements and degradations relative to the base model, respectively.

%=====================================================================

% \usepackage{multirow}

%--------------------------------------
\begin{table}[t]
\centerline{
\resizebox{1.0\linewidth}{!}{
\setlength{\tabcolsep}{0.2mm}{
\begin{tabular}{c|lccc}
\hline
& \multicolumn{1}{c}{}                                 & FLOPs                                & Params                     & Acc                                           \\
\multirow{-2}{*}{Models}  & \multicolumn{1}{c}{\multirow{-2}{*}{Pruning Method}} & (G)                                  & (M)                        & (\%)                                          \\ \hline
& No token reduction                               & 4.6                                  & 22                         & 79.8$(-)$                                     \\
                          & +DynamicViT [NeurIPS'21]                         & 2.9                                  & 22                         & 79.3$(\textcolor{red}{-0.5})$                                  \\
                          & +IA-RED$^2$ [NeurIPS'21]                         & 3.2                                  & 22                         & 79.1$(\textcolor{red}{-0.7})$                                  \\
                          & +EViT [ICLR'22]                                  & 3.0                                  & 22                         & 79.5$(\textcolor{red}{-0.3})$                                  \\
                          & +ATS [ECCV'22]                                   & 2.9                                  & 22                         & 79.7$(\textcolor{red}{-0.1})$                                  \\
                          & +A-ViT [CVPR'22]                                 & 3.6                                  & 22                         & 78.6$(\textcolor{red}{-1.2})$                                  \\
                          & +Evo-ViT [AAAI'22]                               & 3.0                                  & 22                         & 79.4$(\textcolor{red}{-0.4})$                                  \\
                          & +SPViT [ECCV'22]                                 & 2.7                                  & 22                         & 79.3$(\textcolor{red}{-0.5})$                                  \\
                          & +ToMe [ICLR'23]                                  & 2.7                                  & 22                         & 79.4$(\textcolor{red}{-0.4})$                                  \\
                          & +BAT [CVPR'23]                                   & 3.0                                  & 22                         & 79.6$(\textcolor{red}{-0.2})$                                  \\
                          & +eTPS [CVPR'23]                                  & 3.0                                  & 22                         & 79.7$(\textcolor{red}{-0.1})$                                  \\
                          & +dTPS [CVPR'23]                                  & 3.0                                  & 22                         & 80.1$(\textcolor{Green}{+0.3})$                                  \\
                          & +ToFu [WACV'24]                         & 2.7                                  & 22                         & 79.6$(\textcolor{red}{-0.2})$                                  \\
                          & +Zero-TP [CVPR'24]                         & 2.5                                  & 22                         & 79.1$(\textcolor{red}{-0.7})$                                  \\
                          & +ELSA [CVPR'24]                         & 2.2                                  & 22                         & 79.0$(\textcolor{red}{-0.8})$                                  \\
                          & +MCTF$_{r=16}$ [CVPR'24]                         & 2.6                                  & 22                         & 80.1$(\textcolor{Green}{+0.3})$                                  \\
\multirow{-14}{*}{DeiT-S} &  +  BSPF-ViT (Ours)       & \textbf{2.4} & 22 & \textbf{81.8$(\textcolor{Green}{\textbf{+2.0}})$} \\ \hline
& No token reduction                                   & 1.3                                  & 5                          & 72.2$(-)$                                     \\
                          & +DynamicViT [NeurIPS'21]                        & 0.8                                  & 5                          & 71.4$(\textcolor{red}{-0.8})$                                  \\
                          & +EViT [ICLR'22]                                  & 0.8                                  & 5                          & 71.9$(\textcolor{red}{-0.3})$                                  \\
                          & +A-ViT [CVPR'22]                                 & 0.8                                  & 5                          & 71.0$(\textcolor{red}{-1.2})$                                  \\
                          & +Evo-ViT [AAAI'22]                               & 0.8                                  & 5                          & 72.0$(\textcolor{red}{-0.2})$                                  \\
                          & +SPViT [ECCV'22]                                 & 0.9                                  & 5                          & 72.1$(\textcolor{red}{-0.1})$                                  \\
                          & +ToMe [ICLR'23]                                  & 0.7                                  & 5                          & 71.3$(\textcolor{red}{-0.9})$                                  \\
                          & +BAT [CVPR'23]                                   & 0.8                                  & 5                          & 72.3$(\textcolor{Green}{+0.1})$                                  \\
                          & +eTPS [CVPR'23]                                 & 0.8                                  & 5                          & 72.3$(\textcolor{Green}{+0.1})$                                  \\
                          & +dTPS [CVPR'23]                                  & 0.8                                  & 5                          & 72.9$(\textcolor{Green}{+0.7})$                                  \\
                          & +MCTF [CVPR’24]                             & 0.7                                  & 5                          & 72.7$(\textcolor{Green}{+0.5})$                                  \\
\multirow{-12}{*}{DeiT-T} & + BSPF-ViT (Ours)       & \textbf{0.6} & 5  & \textbf{73.5$(\textcolor{Green}{\textbf{+1.3}})$} \\ \hline
\end{tabular}}}
}
\caption{Comparison of BSPF-ViT against other token reduction methods on ImageNet-1K using DeiT~\cite{touvron2021training} baselines. 
%Our method consistently outperforms existing 1D pruning approaches.
\vspace{-2mm}
}
\label{table: classification}
\end{table}
%--------------------------------------

%=============================TABLE===================================
% \usepackage{multirow}
\begin{table}[t]
\vspace{-2mm}
\centerline{
\resizebox{1.0\linewidth}{!}{
\setlength{\tabcolsep}{0.1mm}{
\begin{tabular}{c|lccc}
\hline
& \multicolumn{1}{c}{}                                 & FLOPs                                & Params                       & Acc                                   \\
\multirow{-2}{*}{Models}                          & \multicolumn{1}{c}{\multirow{-2}{*}{Pruning Method}} & (G)                                  & (M)                          & (\%)                                  \\ \hline
  & No token reduction                                   & 6.1                                  & 21.5                         & 81.7 $(-)$                                  \\
                                                  & +MCTF$_{r=13}$[CVPR’24]                              & 4.2                                  & 21.5                         & 81.8 $(\textcolor{Green}{\uparrow})$                                 \\
\multirow{-3}{*}{T2T-ViT$_t$-14}                     &  +  BSPF-ViT (Ours)           & \textbf{3.6} & 21.5 & \textbf{82.6 }$(\textcolor{Green}{\Uparrow})$ \\ \hline
  & No token reduction                                   & 9.8                                  & 39.2                         & 82.4 $(-)$                                 \\
                                                  & +MCTF$_{r=9}$ [CVPR’24]                               & 6.4                                  & 39.2                         & 82.4 $(-)$                                 \\
\multirow{-3}{*}{T2T-ViT$_t$-19}                     & +  BSPF-ViT (Ours)           & \textbf{5.0} & 39.2 & \textbf{83.1}$(\textcolor{Green}{\Uparrow})$ \\ \hline
 & No token reduction                                   & 6.6                                  & 26.2                         & 83.3 $(-)$                                \\
                                                  & +EViT [ICLR’22]                                  & 4.7                                  & 26.2                         & 83.0 $(\textcolor{red}{\downarrow})$                                 \\
                                                  & +BAT [CVPR’23]                                 & 4.7                                  & 26.2                         & 83.1  $(\textcolor{red}{\downarrow})$                                \\
                                                  & +DynamicViT [NeurIPS’21]                        & 4.6                                  & 26.9                         & 83.0   $(\textcolor{red}{\downarrow})$                               \\
                                                  & +SPViT [ECCV’22]                           & 4.3                                  & 26.2                         & 83.1 $(\textcolor{red}{\downarrow})$                                 \\
                                                  & +MCTF$_{r=12 }$ [CVPR’24]                              & 4.2                                  & 26.2                         & 83.3 $(-)$                                 \\
\multirow{-7}{*}{LV-ViT-S}                        &  + BSPF-ViT (Ours)           & \textbf{3.8} & 26.2 & \textbf{84.1}$(\textcolor{Green}{\Uparrow})$ \\ \hline
\multicolumn{1}{l|}{}                             & No token reduction                                   & 4.1                                  & 26                           & 83.6 $(-)$                                 \\
\multicolumn{1}{l|}{}                             & +MCTF$_{r=9}$ [CVPR’24]                               & 2.6                                  & 26                           & 83.6 $(-)$                                 \\
\multicolumn{1}{l|}{\multirow{-3}{*}{CaFormer-S18}} &  +  BSPF-ViT (Ours)          & \textbf{2.5} & 26   & \textbf{84.2}$(\textcolor{Green}{\Uparrow})$ \\ \hline
\end{tabular}}}
}
\caption{Compared with other pruning methods on ImageNet-1K dataset. BSPF-ViT outperforms the SoTA MCTF~\cite{lee2024multi} by $\sim$ 0.5\%. 
}
\label{table: SoTA pruning method}
\vspace{-2mm}
\end{table}
%=====================================================================

%=============================TABLE===================================
% \usepackage{multirow}
\begin{table}[t]
\centerline{
\resizebox{0.9\linewidth}{!}{
\begin{tabular}{lccc}
\hline
\multicolumn{1}{c}{}                         & FLOPs                      & Params                      & Acc                         \\
\multicolumn{1}{c}{\multirow{-2}{*}{Models}} & (G)                        & (M)                         & (\%)                        \\ \hline
PVT-Small                                    & 3.8                        & 24.5                        & 79.8                        \\
PVT-Medium                                   & 6.7                        & 44.2                        & 81.2                        \\
CoaTMini                                     & 6.8                        & 10.0                        & 80.8                        \\
CoaT-LiteSmall                               & 4.0                        & 20.0                        & 81.9                        \\
Swin-T                                       & 4.5                        & 29.0                        & 81.3                        \\
Swin-S                                       & 8.7                        & 50.0                        & 83.0                        \\
PoolFormer-S36                               & 5.0                        & 31.0                        & 81.4                        \\
PoolFormer-M48                               & 11.6                       & 73.0                        & 82.5                        \\
GC ViT-XT                               & 2.6                       & 20.0                       & 82.0                        \\ 
SMT-S                               & 4.7                       & 20.5                       & 83.7                        \\
TransNeXt-Tiny                               & 5.7                       & 28.2                        & 84.0                        \\
RMT                               & 4.5                       & 27.0                        & 84.1                        \\
 \hline 
CaFormer-S18            &  4.1 &  26.0   &  83.6 \\
 + BSPF-ViT (Ours)                                & 2.5                        & 26.0                          & \textbf{84.2}   $(\textcolor{Green}{\Uparrow})$  \\
\hline
\end{tabular}
}}
\caption{Comparison with other ViTs on ImageNet-1K.
\vspace{-2mm}
}
\label{table: other transformer}
\vspace{-2mm}
\end{table}
%=====================================================================

\textbf{Comparison of token reduction methods.} 
Table~\ref{table: classification} presents a performance comparison between our BSPF-ViT and other token reduction methods. Our approach consistently outperforms others, achieving the highest accuracy with the lowest FLOPs on the DeiT~\cite{touvron2021training} baseline. Notably, BSPF-ViT is the only method to achieve a 1.0\% performance improvement while maintaining minimal FLOPs in both DeiT-T and DeiT-S. Specifically, BSPF-ViT boosts accuracy by 1.3\% and reduces FLOPs by 50\% in DeiT-T. Similarly, in DeiT-S, our method achieves a 2.0\% accuracy increase while reducing FLOPs by 2.2 G. While TPS and MCTF achieved some improvements on DeiT-T, their gains on DeiT-S were limited. In contrast, BSPF-ViT consistently enhances DeiT-S accuracy by 2.0\%, demonstrating its superiority to other 1D token pruning techniques.

\textbf{Comparisons of SoTA ViTs w/wo block-based pruning.}
To further assess BSPF-ViT’s effectiveness, we applied it to additional transformer architectures, as summarized in Table \ref{table: SoTA pruning method}. While MCTF achieves a 31\% speedup without compromising performance in these architectures, BSPF-ViT surpasses MCTF, delivering a 10\% speed increase and a 1\% improvement in accuracy. Notably, when combined with LV-ViT, our BSPF-ViT outperforms all other transformers and token reduction methods in both FLOPs and accuracy. Unlike other methods that degrade LV-ViT's performance, BSPF-ViT actually enhances it. These findings confirm that BSPF-ViT is a robust token reduction method across various ViT architectures.

In Table~\ref{table: other transformer}, we also compared the proposed method applied to the CaFormer-S18 architecture within the MetaFormer framework against other ViT architectures. Our method shows promise, achieving a 10\% improvement in speed and a 0.6\% increase in accuracy in CaFormer-S18. Our approach not only enhances model efficiency but also boosts accuracy, further validating that ours method effectively preserves critical information during pruning.

%=====================================================================
\begin{table}[t]
\centerline{
\resizebox{1.02\linewidth}{!}{
\setlength{\tabcolsep}{0.1mm}
\begin{tabular}{c|c|ccc|ccc}
\hline
 & Latency  & \multicolumn{3}{c|}{Object Det.}             & \multicolumn{3}{c}{Instance Seg.}           \\ \cline{3-8} 
\multirow{-2}{*}{Backbone}      &  (ms)           & AP$^{box}$    & AP$^{box}_{50}$ & AP$^{box}_{75}$ & AP$^{mask}$   & AP$^{mask}_{50}$ & AP$^{mask}_{75}$ \\ \hline
ResNet18                        & 0.71          & 34.0          & 54.0            & 36.7            & 31.2          & 51.0             & 32.7             \\
PoolFormer-S12                  & 1.20          & 37.3          & 59.0            & 40.1            & 34.6          & 55.8             & 36.9             \\
EfficientFormer-L1              & 0.86          & 37.9          & 60.3            & 41.0            & 35.4          & 57.3             & 37.3             \\
RepViT-M1.1                     & 0.78          & 39.8          & 61.9            & 43.5            & 37.2          & 58.8             & 40.1             \\
PoolFormer-S24                  & 1.96          & 40.1          & 62.2            & 43.4            & 37.0          & 59.1             & 39.6             \\
PVT-Small                       & 8.59          & 40.4          & 62.9            & 43.8            & 37.8          & 60.1             & 40.3             \\
EfficientFormer-L3              & 1.98          & 41.4          & 63.9            & 44.7            & 38.1          & 61.0             & 40.4             \\
RepViT-M1.5                     & 1.02          & 41.6          & 63.2            & 45.3            & 38.6          & 60.5             & 41.5             \\
EfficientNet-B0                 & 0.54          & 31.9          & 51.0            & 34.5            & 29.4          & 47.9             & 31.2             \\
EfficientViT-M4                 & 0.33          & 32.8          & 54.4            & 34.5            & 31.0          & 51.2             & 32.2             \\
FastViT-SA12                    & 1.06          & 38.9          & 60.5            & 42.2            & 35.9          & 57.6             & 38.1             \\ \hline
SHViT-S4                        & 0.28          & 39.0          & 61.2            & 41.9            & 35.9          & 57.9             & 37.9             \\
\textbf{+  BSPF-ViT (Ours)} & \textbf{0.28} & \textbf{41.9} & \textbf{64.7}   & \textbf{47.6}   & \textbf{40.6} & \textbf{62.4}    & \textbf{42.8}    \\ \hline
\end{tabular}
}}
\caption{Comparison of object detection and instance segmentation on COCO~\cite{lin2014microsoft} using Mask-RCNN~\cite{he2017mask}. GPU latencies were measured with $512 \times 512$ image crops and batch size $32$. 
\vspace{-2mm}
}
\label{table:Downstream Tasks}
%\vspace{-2mm}
\end{table}
%=====================================================================

%=====================================================================
\begin{table}[t]
\centerline{
\begin{tabular}{l|c|cc}
\hline
Model                    & Similarity Matric                         & GFLOPs                      & Top-1Acc.                    \\ \hline
                         & Cosine similarity & 2.4 & 81.8 \\
                         & Pearson correlation                       & 2.4                         & 81.5                         \\
\multirow{-3}{*}{DeiT-S} & Euclidean Distance                        & 2.4                         & 81.2                         \\ \hline
                         & Cosine similarity & 0.6 & 73.5 \\
                         & Pearson correlation                       & 0.6                         & 73.2                         \\
\multirow{-3}{*}{DeiT-T} & Euclidean Distance                        & 0.6                         & 73.4                         \\ \hline
\end{tabular}
}
\caption{Comparison of similarity metrics on ImageNet-1k. 
\vspace{-2mm}
}
\label{table:similarity matrix}
\vspace{-2mm}
\end{table}

%=====================================================================

%%%%%%%%%%%%%%%%%%%%%%%%%%%%%%%%%%
%%%%%%%%%%%%%%%%%%%%%%%%%%%%%%%%%%
\subsection{Downstream Tasks}

\textbf{Object detection and instance segmentation.} We evaluate our BSPF-ViT on object detection and instance segmentation tasks by integrating it into SHViT to test its generalizability. Following \cite{li2023rethinking}, we incorporated the pruned SHViT into the Mask R-CNN framework \cite{he2017mask} and performed experiments on the MS COCO dataset \cite{lin2014microsoft}. As shown in Table \ref{table:Downstream Tasks}, our method outperforms competing models in both APbox and APmask, even with similar model sizes. In particular, it surpasses the RepViT-M1.1 backbone by 2.1 APbox and 3.4 APmask. Compared to SHViT, the pruned SHViT delivers a comparable APbox and a higher APmask, demonstrating the significant advantages and versatility of BSPF-ViT in high-resolution vision tasks. These results clearly highlight the superiority of \textbf{BSPF-ViT} in transferring to downstream vision tasks.

\subsection{Ablation Study}

\textbf{Pruning decisions with and w/o neighboring tokens.} We hypothesize that in ViT tasks, a correlation exists between adjacent tokens because the image is divided into fixed-size tokens fed into the self-attention mechanism. In Table \ref{table:pruning decisions}, we examine the effects of different pruning decisions on our proposed BSPF-ViT across four architectures: DeiT-S, DeiT-T, T2T-ViT, and CaFormer-S18. The results indicate that using the comprehensive expressiveness of neighboring tokens for pruning decisions yields better outcomes than relying on a single token alone. This observation supports our hypothesis that there is indeed a correlation between tokens, and therefore, neighboring tokens should be considered simultaneously when making pruning decisions.

\textbf{Pruning ratio.}
Table \ref{table:reduction ratio} presents the ablation study on different pruning ratios. When the pruning ratio is 0.9 or 0.8, the model maintains stable accuracy (DeiT-S: 81.8\%, DeiT-T: 73.5\%) but with significantly higher computational cost. In contrast, a lower pruning ratio of 0.4 leads to a noticeable drop in accuracy (DeiT-S: 81.0\%, DeiT-T: 72.8\%) due to excessive information loss. Among these, 0.5 provides the best balance, retaining important information while effectively reducing FLOPs (DeiT-S: 2.4G, DeiT-T: 0.6G) without sacrificing accuracy. Then, we choose 0.5 as the optimal pruning ratio for efficiency and performance.

\textbf{Similarity matrix.} We apply BSPF-ViT to the DeiT-S and DeiT-T models, comparing various similarity measures. Results in Table~\ref{table:similarity matrix} clearly show that cosine similarity outperforms Euclidean distance and Pearson correlation. Euclidean distance focuses on absolute differences, which may ignore important similarities in the deep learning feature space. Pearson correlation measures linear relationships, but in high-dimensional spaces, the relationships between features may not be linear, making it less effective than cosine similarity in capturing complex similarities.

\textbf{Ablations studies on base architectures.} In Fig.~\ref{fig:pruning}, we replace traditional attention with block attention to reduce computational complexity. Previously, the Windows attention mechanism  was used for a similar purpose. Table~\ref{table:difference base attention} compares these two attention mechanisms of {\em chunk attention} and {\em windows attention} with vs. w/o BSPF-ViT. Methods without token reduction tend to have lower accuracy. Both methods focus only on elements near the diagonal, ignoring correlations between distant tokens. By incorporating our method, we can preserve important information, improving accuracy and efficiency. Additionally, block attention is more effective than window attention because we align the BSPF-ViT size with the block size, allowing the model to process attention and perform BSPF-ViT in parallel, further enhancing efficiency.

%=====================================================================
\begin{table}[t]
\centerline{
\resizebox{1.0\linewidth}{!}{
\begin{tabular}{lll|lll}
\hline
\multicolumn{1}{c}{}                                  & \multicolumn{1}{c}{FLOPs} & Acc  & \multicolumn{1}{c}{}                                        & \multicolumn{1}{c}{FLOPs}   & Acc                          \\
\multicolumn{1}{c}{\multirow{-2}{*}{Method}}          & \multicolumn{1}{c}{(G)}   & (\%) & \multicolumn{1}{c}{\multirow{-2}{*}{Method}}                & \multicolumn{1}{c}{(G)}     & (\%)                         \\ \hline
\multicolumn{1}{l|}{DeiT-S}   & 4.6                       & 79.8 & \multicolumn{1}{l|}{T2T-ViT$_t$-14} & 6.1                         & 81.7                         \\ \hline
\multicolumn{1}{l|}{one token}                        & 2.4                       & 81.4 & \multicolumn{1}{l|}{one token}                              & 3.6                        & 82.1                         \\
 
\multicolumn{1}{l|}{adjacent} & 2.4                       & 81.8 & \multicolumn{1}{l|}{adjacent}       & 3.6 & 82.6 \\ \hline
 
\multicolumn{1}{l|}{DeiT-T}   & 1.3                       & 72.2 & \multicolumn{1}{l|}{CaFormer-S18}   & 4.1                         & 83.6                         \\ \hline
\multicolumn{1}{l|}{one token}                        & 0.6                       & 73.2 & \multicolumn{1}{l|}{one token}                              & 2.5                         & 84.0                         \\ 
\multicolumn{1}{l|}{adjacent} & 0.6                       & 73.5 & \multicolumn{1}{l|}{adjacent}       & 2.5 & 84.2 \\ \hline
\end{tabular}}
}
\caption{Ablation study of pruning decisions w/wo neighboring tokens evaluated on the ImageNet-1K dataset for evaluation. 
\vspace{-2mm}
}
\label{table:pruning decisions}
\vspace{-2mm}
\end{table}
%=====================================================================

%=====================================================================
\begin{table}[t]
\centerline{
\resizebox{1.0\linewidth}{!}{
\begin{tabular}{lc|cccccc}
\hline
\multicolumn{2}{l|}{Pruning ratio}                & 0.4  & 0.5           & 0.6  & 0.7  & 0.8  & 0.9  \\ \hline
\multicolumn{1}{l|}{\multirow{2}{*}{DeiT-S}} & ACC (\%)  & 81.0 & \textbf{81.8} & 81.7 & 81.7 & 81.7 & 81.8 \\
\multicolumn{1}{l|}{}                        & FLPOS (G) & 2.3  & \textbf{2.4}  & 2.7  & 3.5  & 3.7  & 4.0  \\ \hline
\multicolumn{1}{l|}{\multirow{2}{*}{DeiT-T}} & ACC (\%)  & 72.8 & \textbf{73.5} & 73.4 & 73.4 & 73.5 & 73.5 \\
\multicolumn{1}{l|}{}                        & FLPOS (G) & 0.5  & \textbf{0.6}  & 0.6  & 0.7  & 0.9  & 1.1  \\ \hline
\end{tabular}}
}
\caption{Ablation study of the effect of pruning scale ses the
ImageNet-1k dataset for evaluation. 
\vspace{-2mm}
}
\label{table:reduction ratio}
\vspace{-2mm}
\end{table}
%=====================================================================

%================================================
\begin{table}[t]
\centerline{
\resizebox{1.0\linewidth}{!}{
\begin{tabular}{lccc}
\hline
Base Architecture                  & Pruning method         & GFLOPs       & Top-1Acc.(\%) \\ \hline
\multirow{2}{*}{Chunk attention}   & No token reduction     & 2.4          & 81.4          \\
                                   & \textbf{BSPF-ViT} & \textbf{2.4} & \textbf{81.8} ($\Uparrow$)\\ \hline
\multirow{2}{*}{Window attention} & No token reduction     & 3.6          & 80.6          \\
                                   & BSPF-ViT          & 3.6          & 81.0 ($\Uparrow$)         \\ \hline
\end{tabular}}
\vspace{-2mm}
}
\caption{Comparison of base architectures on ImageNet-1k. 
\vspace{-2mm}
}
\label{table:difference base attention}
\vspace{-2mm}
\end{table}
%==============Experiment Results===============

% \textred{describe Table~\ref{table:reduction ratio}.}

%=====================================================================

% \begin{algorithm}
% \caption{SARAS-Net for change detection}\label{euclid}
% \hspace*{\algorithmicindent} \textbf{Input:}  Two temporal remote sensing images ($X$, $Y$) \\
% \hspace*{\algorithmicindent} \textbf{Output:}  Change map $M$
% \begin{algorithmic}[1]
% \State  \textit{ \textbf{Step1}: Feature Extraction}\\
% $F_X$ = $CNN$($X$);   $F_Y$ = $CNN$($Y$);
% \State \textit{ \textbf{Step2}: Relation-aware}
% \For{layer $n$ }
%     \State $ ({\bar F}_X^n, \bar F_Y^n)$ = Relation-Aware-module($F_X^n$,$F_Y^n$); 
%     \State $ D_n=abs({\bar F}_X^n- \bar F_Y^n)$;  
% \EndFor
% \State \textit{\textbf{Step3}: Scale-aware adn Cross-Transformer}
% \For{layer $n$ }
% { \\ \  \ \quad ${U}_n$ = Scale-Aware-Attention($D_{n}$); 
%     \For{layer $m$ ($m$ $\neq$ $n$) }{ \\
%         \qquad \ $D_m^n$ = Channel-wise($U_n$, Resize($D_m$));
%     }
%     \EndFor
%     $S_{n}$ =  CTB($D_1^n$, $D_2^n$, $D_3^n$, $D_4^n$); 
% }
% \EndFor

% \State \textit{ \textbf{Step4}: Change Map Generation}
% \State 
% \quad $P$ = Softmax($G$ ($S_{1}$\copyright $S_{2}$\copyright $S_{3}$\copyright $S_{4}$));
% \State  \Return P 
% \end{algorithmic}

% \end{algorithm}

%%%%%%%%%%%%%%%%%%%%%%%%%%%%%%%%%
%%%%%%%%%%%%%%%%%%%%%%%%%%%%%%%%%
%%%%%%%%%%%%%%%%%%%%%%%%%%%%%%%%%
\section{Conclusions}

This paper presents a novel block-based token pruning method for efficient ViTs, balancing accuracy and computational efficiency. By leveraging the symmetry between query and key, we maintain a symmetric attention matrix, simplifying computation during pruning. Our method evaluates token importance in conjunction with neighboring tokens, enabling more precise pruning. Through similarity-based fusion, pruned information is restored into retained tokens, which enhances data integrity and model accuracy. We improve the DeiT-S model’s accuracy by 2.0\% and reduce FLOPs by 2.2G, with applications to downstream tasks demonstrating its versatility. Unlike 1D pruning methods, our method captures broader token interactions and supports parallel block processing, speeding up inference. Future work will explore adaptive block sizes and flexible pruning strategies to reduce computational costs further.

{
    \small
    \bibliographystyle{ieeenat_fullname}
    \bibliography{main}
}


\end{document}